\documentclass[10pt,twocolumn,letterpaper]{article}

\usepackage[article]{cvpr}

\usepackage{graphicx}
\usepackage{amsmath}
\usepackage{amssymb}
\usepackage{booktabs}
\usepackage{ctable}
\usepackage{multirow}
\usepackage{array}
\usepackage{colortbl}
\usepackage{float}
\usepackage[bottom]{footmisc}

\definecolor{cvprblue}{rgb}{0.21,0.49,0.74}
\usepackage[pagebackref,breaklinks,colorlinks,citecolor=cvprblue]{hyperref}

\def\paperID{}
\def\confName{CVPR}
\def\confYear{}

\begin{document}
\renewcommand{\thefootnote}{\fnsymbol{footnote}}

\title{Preparation of Fractal-Inspired Computational Architectures for \\ Advanced Large Language Model Analysis\protect}

\author{
\begin{tabular*}{\textwidth}{@{\extracolsep{\fill}}ccc}
Yash Mittal\textsuperscript{*} & Dmitry Ignatov\textsuperscript{*} & Radu Timofte \\[0.3em]
\footnotesize\texttt{\mbox{yash.mittal@stud-mail.uni-wuerzburg.de}} &
\footnotesize\texttt{\mbox{dmytro.ignatov@uni-wuerzburg.de}} &
\footnotesize\texttt{\mbox{radu.timofte@uni-wuerzburg.de}}
\end{tabular*}\\[0.5em]
\small Computer Vision Lab, CAIDAS, University of W\"urzburg, Germany
}
\date{}
\maketitle
\footnotetext[1]{Corresponding authors}

\begin{abstract}
This paper proposes a framework called FractalNet based on fractal design principles that automatically generate and evaluate convolutional neural network (CNN) architectures by using template patterns. The main idea of the framework is not to depend on costly Neural Architecture Search (NAS) operations but instead to explore a well-structured architecture space which is specified by recursive fractal templates that change major parameters such as fractal depth, column width, and layer configurations. The framework comprises three major parts: a generator to create candidate architectures by performing controlled permutations of convolutional, normalization, activation, and dropout layers; a fractal template module to impose recursive multi-path structural patterns; and a runner module to deal with model training, evaluation, and logging. Through the use of this system, over 1,200 different CNN architectures were automatically created and tested on the CIFAR-10 image classification dataset. Training was done in PyTorch by applying stochastic gradient descent together with Automatic Mixed Precision (AMP) and gradient checkpointing to speed up computations. The results of the experiments indicate that fractal-based architectures have stable training characteristics and are capable of achieving competitive results, i.e. an average validation accuracy of around $\sim$60--70\% and maximum performance of 80.18\% when training for just five epochs. The findings indicate that recursive fractal structures are a very good means of finding the right balance between network depth and width, while also supporting large-scale architecture exploration. The new framework presents a less resource-demanding and more understandable way of performing extensive automated architecture experiments.
\vspace{-0.55cm}
\end{abstract}

\section{Introduction}
\label{sec:intro}
Deep learning is responsible for the impressive results in many fields, such as computer vision, natural language processing, and multimodal learning. One of the main reasons for these progresses is the designing of brain-like structures that are highly performative. In the past, a lot of the architectures that have had a great impact to this day were designed manually by very experienced people through their understanding of the scripts and lots of testing. The problem with designing architectures by hand is that it takes a lot of time and also in many cases, requires considerable amount of computing power which is why people started to think about using automated architecture design methods.

\begin{figure}[!ht]
    \centering
    \includegraphics[width=\columnwidth]{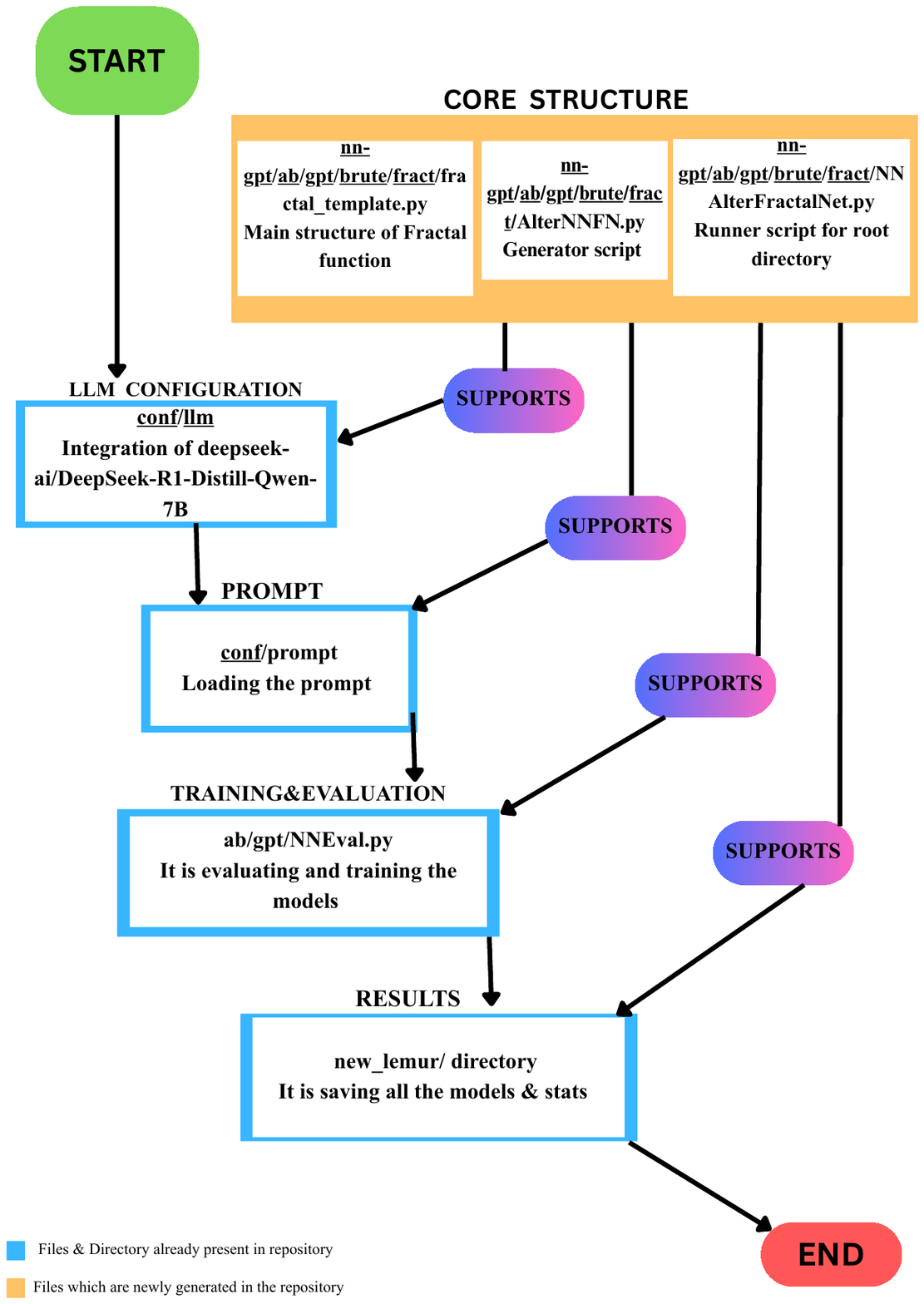}
    \caption{Workflow of FractalNet}
    \label{fig:workflow-fractalnet}
\end{figure}

Quite a few NAS and AutoML frameworks have appeared on the market as excellent aids in fully automatic discovery of top performing networks. Various NAS variants including those based on reinforcement learning, and new differentiable ones like ENAS or DARTS have achieved remarkable success on standard image datasets like CIFAR-10 and ImageNet. However, almost every NAS technique needs a huge amount of processing power, at least thousands of GPU hours, and the newly invented architectures are typically so complicated that even experts find them hard to work with, which is a huge barrier for a thorough architectural study and carrying out controlled experiments.

Another approach at least superficially opposite to this, looks into the use of well-organized structural architectural predictions for the purpose motivating network design. Fractal neural networks are one of the examples of this kind of method, where by using recursive self-similarity patterns, deep networks with multiple paths of computation are created. The very first FractalNet showed that an extremely deep network could be trained very well even without residual connections by using fractal connectivity and implicit ensemble behavior of multiple paths. These recursive patterns facilitate the reuse of features and the flow of gradients in a stable manner while at the same time making the architectural design more transparent.

Meanwhile, other studies have looked into the idea of using large language models (LLMs) for AutoML. LLM-powered solutions can create different architecture setups, propose hyperparameters, or even help you locate your search path just by describing what you want in natural language. Sure, these methods further automate the design of neural architecture, but basically, they are at the configuration level only and generally do not come with explicit structural constraints that help in architecture generation.

The advances in automated architecture search and structured neural design are excellent. But, even then, not much research has been done on mixing structured architectural priors with scalable automated generation pipelines. For instance, fractal architectures are an excellent design space because of their recursive structure, but huge numbers of fractal variants have not been systematically explored. A system that can automatically make a big variety of fractal-based architectures at the same time still keeping them understandable and efficient would be very useful for large-scale architecture experiments.

Therefore, this article proposes FractalNet, a template-driven tool for automatically designing and testing fractal- inspired CNN architectures. The method adopts recursive fractal templates to specify a structured architecture search space, and it automatically produces candidate models by changing parameters such as fractal depth, column width, and layer configuration. The tool combines three key parts: a producer that creates architectural designs, a fractal template module to assemble recursive network structures, and a runner for training, evaluation, and logging performance.

In this way, over 1,200 unique CNN architectures are automatically created and tested on the CIFAR-10 dataset. The experiments look at how structural parameters like fractal depth and column width affect the model's performance, convergence patterns and computational efficiency. The findings show that fractal-inspired architectures not only train very well but also allow for scalable architecture exploration with relatively modest computational costs.

This research's chief contributions are highlighted below:
\begin{itemize}
\item A framework grounded in templates is proposed by us for the automated generation of fractal-inspired convolutional neural network architectures.
\item A structured architecture search space is introduced by us based on fractal depth, column width and configurable layer sequences.
\item An automated training and evaluation pipeline is developed by us which is capable of generating and benchmarking more than 1, 200 neural network variants.
\item An empirical analysis of fractal architecture configurations on the CIFAR-10 dataset, their performance, convergence behavior, and computational efficiency is provided by us.
\end{itemize}

Put simply, this research suggests that fractal-inspired architectural design can be a powerful and interpretable basis of automated neural architecture exploration, providing a scalable substitute for computationally expensive architecture search methods.

\section{Related Work}
\label{sec:Related}
There is no doubt that automated neural architecture design has become a major focus in deep learning research. Below, we discuss three areas that are highly relevant to our work: neural architecture search (NAS) methods, fractal neural network architectures, and recent developments in incorporating large language models into AutoML pipelines.
\subsection{Neural Architecture Search}
Neural Architecture Search (NAS) is a technique that aims at automatically discovering top-performing neural network architectures without human intervention in design. Most of the initial NAS methods used reinforcement learning or evolutionary algorithms to generate large populations of candidate architectures for further evaluation. To give an example, in their work, Krizhevsky et al.~\cite{Krizhevsky2012} highlighted the power of deep convolutional neural networks for large-scale image classification, which opened up new avenues for research in automated architecture design. Following works proposed more efficient NAS strategies including weight sharing and differentiable search techniques. Among the prominent ones are Efficient Neural Architecture Search (ENAS) and differentiable architecture search frameworks that drastically cut down the computational resources needed for architecture discovery.
\subsection{Fractal Neural Networks}
Fractal neural networks use recursive self-similar structures to create deep architectures. The FractalNet framework by Larsson et al.~\cite{larsson2017fractalnetultradeepneuralnetworks} was a breakthrough by showing that very deep networks can be built without residual connections if you use multiple parallel paths of different depths. This recursive pattern enhances the gradient flow and makes the network more inclined to reuse features during training. On the other hand, most of the work done until now is centered around fixed fractal architectures and not on exploring various fractal configurations systematically.
\begin{figure}[ht]
    \centering
    \includegraphics[width=\columnwidth]{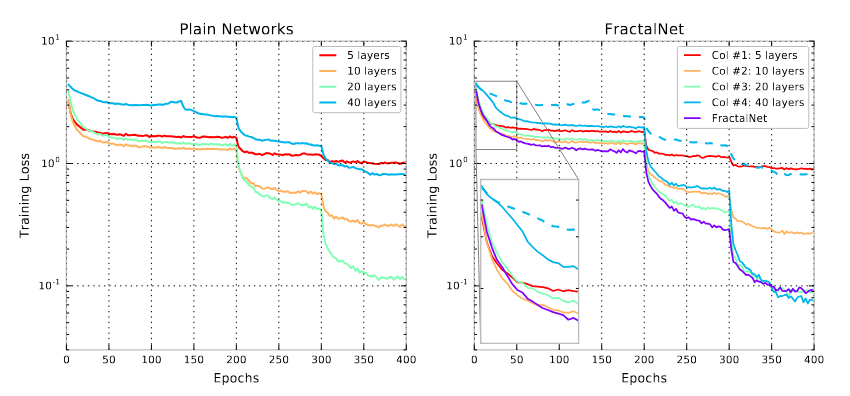}
    \caption{Comparison of training loss progression across epochs between Plain Networks and FractalNet \protect\cite{larsson2017fractalnetultradeepneuralnetworks}.}
    \label{fig:comparison_graph1}
\end{figure}

\subsection{LLM-Assisted AutoML}
Some of the recent research has examined how large language models (LLMs) could be used to facilitate automatic machine learning processes. A few systems like NNGPT and HPGPT~\cite{ABrain.NNGPT, ABrain.HPGPT, ABrain.NN-Dataset,ABrain.NN-Lite} exploit LLMs for generating architecture configurations or for giving hyperparameter suggestions. Even though these methods enhance automation, they mostly work at the configuration level and do not set any clear structural limitations on the architectures they generate.
\subsection{Position of This Work}
This study highlights automated building of neural network architectures in a structured search space that is inspired by fractal design principles. Our method integrates recursive fractal templates with automatic generation and assessment, so it allows for extensive investigation of fractal-like CNN architectures without giving up on maintaining interpretable structural patterns.

\section{Methodology}
\label{sec:Methodology}
Recent advances in the application of large language models (LLMs) across diverse domains~\cite{ABrain.HPGPT,Gado2025llm,Rupani2025llm,ABrain.NN-RAG}, combined with prior architectural synthesis work within the NNGPT framework~\cite{ABrain.HPGPT,ABrain.NN-Captioning_2025,ABrain.Prompt,ABrain.Transform,ABrain.Architect,ABrain.CV_Channel,ABrain.Feedback_Memory,ABrain.Delta}, motivate the present work. Building upon the LEMUR dataset of high-capacity and resource-efficient neural network models~\cite{ABrain.NN-Dataset,ABrain.LEMUR2,ABrain.NN-Lite,ABrain.MobileAgeNet,ABrain.MobileDenoising}, we propose an automated pipeline for the design, training, and evaluation of convolutional neural networks grounded in a principled fractal architectural template within the NNGPT framework~\cite{ABrain.NNGPT}. The pipeline comprises three tightly integrated components: a \textit{Generator}, a \textit{Fractal Template}, and a \textit{Runner}. Together, these components enable systematic exploration of a large space of neural network architectures with minimal human intervention, supporting reproducible and scalable evaluation across diverse design configurations.
\subsection{Architecture Search Space}
The architecture search space is determined mainly by two structural parameters: \textit{fractal depth} ($N$) and \textit{column width} ($num\_column$). While fractal depth sets how many times the network is built recursively (or how "deep" the network is), column width tells how many different parallel paths are in one fractal block at the same time. And these parameters make it possible for the framework to look into networks that differ in both depth and width but still have a common structural pattern.
Officially, a generated network design can be defined as [A=(N, num\_column, L)] where $N$ is the fractal depth, $num\_column$ is the number of parallel columns, and $L$ is the list of layer operations within each block. The set of layer operations includes convolution, normalization, activation, and dropout.
\subsection{Generator Module}
The Generator module is the one which creates candidate neural network configurations or simply architectures. It methodically creates architectures by going through combinations of different structural parameters as well as layer configurations. The generator changes various parts of the architecture such as convolutional kernel sizes, activation functions, normalization levels, and dropout rates. For each set of parameters, the generator outputs a candidate architecture specification which is then realized by the fractal template module. During generation, incompatible configurations like impossible sequences of layers or unstable structural combinations are eliminated to ensure that the resulting architectures are suitable for the training engine.

\begin{table}[H]
\centering
\begin{tabular}{ll}
\hline
Parameter & Values \\
\hline
Kernel Size & {3x3} \\
Activation & {ReLU} \\
Normalization & {BatchNorm} \\
Dropout & {0.2} \\
Depth $N$ & {1--6} \\
Columns & {1--8} \\
\hline
\end{tabular}
\caption{Generator Search Space}
\end{table}

\subsection{Fractal Template Construction}
The Fractal Template creates the network architecture with fractal blocks at multiple levels. Each block has several parallel ways, which mainly only differ in their depth levels but otherwise have very similar structural patterns. This design feature allows a single network to have both shallow and deep features representations.
The fractal structure follows a recursive definition \cite{Huang2016a}:
\[
F(N) =
\begin{cases}
\text{ConvBlock}, & \text{if } N = 1 \\
\text{Concat}(F(N-1), \text{ConvBlock}), & \text{if } N > 1
\end{cases}
\]

Where $\text{ConvBlock}$ is a convolutional unit consisting of convolution, normalization, activation, and optional dropout layers.
The recursive nature of the fractal construction enables the creation of multiple pathways of different lengths, which in turn allows the network to combine features from various levels of hierarchy. Besides, fractal connectivity facilitates better gradient flow and leads to feature reuse during training, which in turn makes it possible to reliably optimize even quite deep networks.
\subsection{Runner Module}
The Runner module takes care of the training and testing of the generated architectures. It prepares datasets, sets up initial model weights, and executes the training process under a fixed configuration so that the performance of different architectures can be compared fairly. Training is done in PyTorch with Stochastic Gradient Descent (SGD). To save on computation time and memory, the system makes use of Automatic Mixed Precision (AMP) along with gradient check-pointing. These techniques enable the system to handle the training of many architectures even with limited hardware resources. The runner monitors and records during training core metrics such as training loss, validation accuracy, GPU memory usage, and training time. All data is automatically saved and archived for future reference, thus allowing a structured comparison among various fractal configurations. To sum up, the new approach presents an extendable, reproducible method chain for the design and assessment of fractal-inspired neural network architectures through automatic generation and evaluation.

\section{Experiments}
This section details the experimental setup used to evaluate the proposed FractalNet architecture generation framework across three dimensions: predictive performance, consistency across generated architectures, and computational efficiency. All experiments employ advanced data augmentation techniques following~\cite{Aboudeshish2025augmentation}.
\subsection{Dataset}
To conduct the experiments, we use the CIFAR-10 dataset ~\cite{KaggleCIFAR10} which is considered one of the standards for evaluating image classification performance. It contains 60, 000 small $32\times32$ RGB images that are classified into ten different categories such as airplane, automobile, bird, cat, deer, dog, frog, horse, ship, and truck.

The dataset has 50, 000 images for training and 10, 000 for testing. In training, 45, 000 out of 50, 000 images are utilized to train the model, and 5, 000 images are kept aside for validation. The test set is only used to report the final results.
\subsection{Data Preprocessing and Augmentation}
To obtain better generalization results, we employ the typical data preprocessing and augmentation methods that are used in training. They are:

\begin{itemize}
\item Random horizontal flipping
\item Random cropping with padding
\item Pixel value normalization
\end{itemize}

These modifications make models capable of learning stronger visual representations besides decreasing the chance of overfitting.
\subsection{Training Configuration}
In order to maintain the fairness of the comparison, all the architectures that we generate are trained with the identical set of tuning parameters. Using PyTorch framework, the training is performed with the help of the Stochastic Gradient Descent (SGD) optimizer.
The key training parameters are given in Table~\ref{tab:hyperparameters}.

\begin{table}[h!]
\centering
\begin{tabular}{|l|c|}
\hline
\textbf{Hyperparameter} & \textbf{Value} \\ \hline
Learning Rate           & 0.01           \\ \hline
Batch Size              & 16             \\ \hline
Dropout                 & 0.2            \\ \hline
Momentum                & 0.9            \\ \hline
Transformation          & norm\_flip     \\ \hline
Epochs                  & 5              \\ \hline
\end{tabular}
\caption{Training configuration and hyperparameters used for all FractalNet model variants.}
\label{tab:hyperparameters}
\end{table}

To make the computation faster and allow the training of many models even with limited hardware resources, the training pipeline combines Automatic Mixed Precision (AMP) and gradient checkpointing. These methods drastically lower the GPU memory requirements and give the framework the capability to test a wide range of architectures within the constraints of real resources.
\subsection{Evaluation Protocol}
Training of computer vision models are performed using the AI $\text{Linux}$ docker image $\texttt{abrainone/ai-linux}$\footnote{AI Linux: \scriptsize \url{https://hub.docker.com/r/abrainone/ai-linux}} on $\text{NVIDIA GeForce RTX 3090/4090}$ $24\text{G}$ GPUs of the $\text{CVL Kubernetes}$ cluster at the University of Würzburg and a dedicated workstation.

The evaluation pipeline utilizes a standardized protocol that is uniformly applied to all generated architectures.First, any generated model is automatically created and checked to be sure it fits the PyTorch training framework. Moreover, each architecture is given the same dataset, optimizer parameters, and training schedule. Such a controlled experimental setup makes it possible to fairly compare models that vary solely in their structural configuration.
At last, performance indicators are gathered after each epoch. The gathered indicators cover validation accuracy, training loss, GPU memory usage, and training time. All these parameters are recorded automatically for all models and afterwards used to see how fractal depth and column width affect network performance.Through this evaluation pipeline, the framework is capable of creating, training, and evaluating more than 1, 200 unique neural network architectures in a reliable and reproducible manner.

\section{Results and Discussion}
Experimental results obtained from the training and the evaluation of the automatically generated FractalNet architectures are presented in this section. The objective of this study is to analyze the performances, the convergence behaviors, and the computing characteristics of the generated models. Besides, the study also aims at understanding the effects of structural parameters such as fractal depth and column width on the performance of the network.
\subsection{Overall Performance}
With the use of the proposed automated pipeline, over 1,200 neural network architectures have been generated and evaluated on the CIFAR-10 dataset. To ensure consistency in the comparison, all the architectures were trained using the experimental setup and procedure mentioned in the previous section.
Table~\ref{tab: Overall training} displays the summary of the overall results obtained from different architecture generations.

\begin{table}[h!]
\centering
\begin{tabular}{l l}
\hline
\textbf{Metric} & \textbf{Value} \\
\hline
Average validation accuracy & $\sim$60--70\% \\
Top validation accuracy & $\sim$80.18\% \\
Mean training time per epoch & $\sim$5 minutes \\
Mean GPU memory consumption & 4--5~GB \\
Rate of successful training & $\sim$97\% \\
\hline
\end{tabular}
\caption{Overall Training and Validation Statistics}
\label{tab: Overall training}
\end{table}

The sampled architectures show the multifold advantages of fractal-based deep learning models that comprise stable convergence during training, model balancing of complexity with computational resource usage, and competent generalisation performance even with short training cycles.
\subsection{Effect of Fractal Depth and Column Width}
This study focuses mainly on fractal depth ($N$) and column width ($num\_column$) which are two major architectural parameters influencing performance. Fractal depth is the number of reoccurrences of the fractal pattern while column width controls how many parallel paths can run simultaneously.
Experimental observations show that architectures with moderate fractal depth ($N=3$ or $N=4$) and column width values between $3$ and $4$ consistently produce the best validation accuracy.These settings strike a good balance between the network depth and computational demand. Networks that are too shallow will have a very restricted capacity for representation whereas very deep fractal structures could in the initial stages of training cause problems in optimization.
The multiple parallel paths found in fractal blocks result in abundant opportunities for features to be reused among layers. This architectural aspect not only facilitates less gradient degradation but also supports speedier initial stages of learning when juxtaposed with simple sequential convolutional architectures.
\subsection{Training Behavior}
\begin{figure}[H]
    \centering
    \includegraphics[width=\columnwidth]{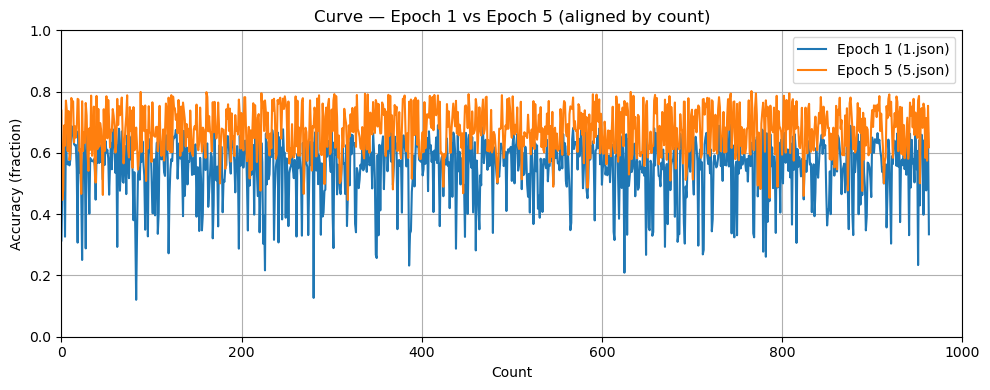}
    \caption{Validation accuracies of all FractalNet models over the first and fifth epochs, showing both initial learning dynamics and final convergence stability\cite{Khalid2023NNStat}}. 
    \label{fig:val_acc_combined}
\end{figure}

Figure~\ref{fig:val_acc_combined} shows the validation accuracy of the generated architectures after the first and the fifth epochs of training. It not only presents how the model initially learns, but also compares the level of training stabilization of the different architectures.

Identifying the change from the first to the fifth epoch reveals that most of the models consistently improve their validation performance. In other words, this strengthens the idea that the generated architectures can successfully capture meaningful features even if training is very limited in time. It is a fact that longer training sessions would lead to higher performance; however, by running these experiments, we have gained valuable knowledge about the early convergence behavior of fractal networks.
\subsection{Computational Efficiency}
Besides classification accuracy, computational efficiency is equally important, especially when it comes to architecture exploration on a very large scale. Errors and training time are significantly cut down by the use of Automatic Mixed Precision (AMP) and gradient checkpointing. These practices made it possible for the framework to even evaluate a big set of models within very limited hardware resources.
However, a majority of the generated architectures will utilize 4 and 5 GB of GPU memory while training and they will need around five minutes for one epoch of training. This is a clear indication that the proposed pipeline can still handle large-scale architecture exploration without requiring massive computational resources.
\subsection{Discussion}
Firstly, one of the major takeaways from the experiments is that fractal-inspired architectures constitute a very rich, flexible yet interpretable design space for automated generation of neural networks. Since fractal blocks are composed in a recursive manner, networks can rely on fractal blocks to be complex enough to have multiple feature extraction pathways that are combined. This way the network optimization process becomes stable, and feature reuse is maximized.
On the contrary to state-of-the-art NAS methods on CIFAR-10 whose reported accuracies are much higher, here the main focus was not to beat those highly optimized NAS models but only to showcase a viable framework for the exploration of neural architectures structured in an automated manner. The fact that fractal templates can be considered as a useful and affordable way of generating diverse neural network designs which can then be systematically analyzed and compared is what reflects the results.

\section{Conclusion}
This article introduced fractalNet, a framework based on templates at the conceptual level to implement the automated generation and evaluation of fractal-inspired convolutional neural network architectures. The combination of fractal recursive design with an automated pipeline allows the framework to perform systematic exploration of architectures with varying depth, width, and layer configurations.

The conducted experiments on the CIFAR-10 dataset indicate that the generated architectures exhibit stable training behavior and achieve good performance even with a short training regime, at the same time requiring only moderate computational resources. In fact, models with a well-balanced fractal depth and column width deliver the best balance between accuracy and computational costs.

In this research, short training schedules and only one dataset were employed. We will extend our work by using longer training sessions, better optimization methods, and evaluating larger and more complex datasets. Altogether, the results demonstrate that fractal-based templates are a very powerful and at the same time interpretable technique for large-scale neural architecture exploration.
\vspace{0.2cm}

\noindent\textbf{Acknowledgments.}
This work was partially supported by the Alexander von Humboldt Foundation.

{
		\small
		\bibliographystyle{ieeenat_fullname}
		\bibliography{bibmain}

@String(NIPS= {Adv. Neural Inform. Process. Syst.})

@String(ICPR = {Int. Conf. Pattern Recog.})

@String(ICLR = {Int. Conf. Learn. Represent.})

@String(CVPRW= {IEEE Conf. Comput. Vis. Pattern Recog. Worksh.})

@String(NIPS  = {NeurIPS})

@String(ICPR  = {ICPR})

@String(ICLR  = {ICLR})

@String(CVPRW= {CVPRW})

@misc{Khalid2023NNStat,
  author       = {W. Khalid},
  title        = {{NN-Stat: Neural Network Statistical Analysis Toolkit}},
  howpublished = {\url{https://github.com/ABrain-One/nn-stat}},
  year         = {2023},
  note         = {GitHub repository}
}

@article{Huang2016a,
  author = {Gao Huang and Zhuang Liu and Kilian Q. Weinberger},
  title = {{Densely connected convolutional networks}},
  journal = {arXiv preprint arXiv:1608.06993},
  year = {2016},
  note = {Published as a conference paper at ICLR 2017}
}

@inproceedings{Krizhevsky2012,
  author = {Alex Krizhevsky and Ilya Sutskever and Geoffrey E. Hinton},
  title = {{ImageNet classification with deep convolutional neural networks}},
  booktitle = {Advances in Neural Information Processing Systems 25 (NIPS)},
  year = {2012}
}

@misc{KaggleCIFAR10,
author = {{Kaggle}},
title = {{CIFAR-10 - Classification in Tiny Images}},
howpublished = {Kaggle Competition},
year = {2017},
url = {https://www.kaggle.com/c/cifar-10/},
note = {Accessed: October 2025}
}

@article{larsson2017fractalnetultradeepneuralnetworks,
  title={FractalNet: Ultra-Deep Neural Networks without Residuals},
  author={Larsson, Gustav and Maire, Michael and Shakhnarovich, Gregory},
  journal={arXiv preprint arXiv:1605.07648},
  year={2017}
}

@InProceedings{ABrain.NNGPT,
	title = {{NNGPT}: Rethinking {AutoML} with Large Language Models},
	author = {Kochnev, Roman and Khalid, Waleed and Uzun, Tolgay Atinc and Zhang, Xi and Dhameliya, Yashkumar Sanjaybhai and Qin, Furui and Vysyaraju, Chandini and Duvvuri, Raghuvir and Goyal, Avi and Ignatov, Dmitry and Timofte, Radu},
	booktitle={Proceedings of the IEEE/CVF Conference on Computer Vision and Pattern Recognition Workshops (CVPRW)},	
	year={2026},
         note={to appear}
}

@article{ABrain.NN-Captioning_2025,
    title = {{LLM} as a Neural Architect: Controlled Generation of Image Captioning Models Under Strict {API} Contracts},
	author = {Krunal Jesani and Dmitry Ignatov and Radu Timofte},
	journal = {arXiv preprint},
  	volume  = {arXiv:2512.14706},
  	url = {https://arxiv.org/abs/2512.14706},
	year={2025}
}

@InProceedings{ABrain.HPGPT,
	title={Optuna vs Code Llama: Are {LLMs} a New Paradigm for Hyperparameter Tuning?},
	author={Kochnev, Roman and Goodarzi, Arash Torabi and Bentyn, Zofia Antonina and Ignatov, Dmitry and Timofte, Radu},
	booktitle={Proceedings of the IEEE/CVF International Conference on Computer Vision Workshops (ICCVW)},
	url={https://openaccess.thecvf.com/content/ICCV2025W/AIM/papers/Kochnev_Optuna_vs_Code_Llama_Are_LLMs_a_New_Paradigm_for_ICCVW_2025_paper.pdf},
	pages = {5664--5674},
	year={2025}
}

@article{ABrain.NN-Dataset,
	title={{LEMUR} Neural Network Dataset: Towards Seamless {AutoML}},
	author={Goodarzi, Arash Torabi and Kochnev, Roman and Khalid, Waleed and Qin, Furui and Uzun, Tolgay Atinc and Dhameliya, Yashkumar Sanjaybhai and Kathiriya, Yash Kanubhai and Bentyn, Zofia Antonina and Ignatov, Dmitry and Timofte, Radu},	
	primaryClass = {cs.LG},	
	journal = {arXiv preprint},
  	volume  = {arXiv:2504.10552},
  	url = {https://arxiv.org/abs/2504.10552},
	year={2025}
}

@InProceedings{ABrain.LEMUR2,
	title={{LEMUR} 2: Unlocking Neural Network Diversity for {AI}},
	 author={Uzun, Tolgay Atinc and Khalid, Waleed and Din, Saif U and Mulukuledu, Sai Revanth and Singh, Akashdeep and Vysyaraju, Chandini and Duvvuri, Raghuvir and Goyal, Avi and Lukhi, Yashkumar Rajeshbhai and Hussain, Ahsan and Jesani, Krunal and Shrestha, Usha and Mittal, Yash and Kochnev, Roman and Kadam, Pritam and Ikram, Mohsin and Moradiya, Harsh Rameshbhai and Arslanian, Alice and Ignatov, Dmitry and Timofte, Radu},
	booktitle={Proceedings of the IEEE/CVF Conference on Computer Vision and Pattern Recognition Workshops (CVPRW)},	
	year={2026},
         note={to appear}
}

@article{ABrain.Transform,
	title={From Brute Force to Semantic Insight: Performance-Guided Data Transformation Design with {LLMs}},
	author={Shrestha, Usha and Ignatov, Dmitry and Timofte, Radu},
	journal={arXiv preprint},
	volume  = {arXiv:2601.03808},
	url = {https://arxiv.org/abs/2601.03808},
	year={2026}
}

@InProceedings{ABrain.Architect,
	title={From Memorization to Creativity: {LLM} as a Designer of Novel Neural Architectures},
	author={Khalid, Waleed and Ignatov, Dmitry and Timofte, Radu},
	booktitle={Proceedings of the IEEE/CVF Conference on Computer Vision and Pattern Recognition Workshops (CVPRW)},	
	year={2026},
         note={to appear}
}

@article{ABrain.NN-Lite,
    title = {{AI} on the Edge: An Automated Pipeline for {PyTorch-to-Android} Deployment and Benchmarking},
	author = {Saif U Din and Muhammad Ahsan Hussain and Mohsin Ikram and Dmitry Ignatov and Radu Timofte},
	doi = {10.20944/preprints202511.1831.v1},
	url = {https://doi.org/10.20944/preprints202511.1831.v1},
	year = 2025,
	month = {November},
	publisher = {Preprints},
	journal = {Preprints}
}

@article{Rupani2025llm,
  	title={Exploring the Collaboration Between Vision Models and {LLMs} for Enhanced Image Classification},
	author={Rupani, Bhavya and Ignatov, Dmitry and Timofte, Radu},
	year=2025,
	month = {December},
	publisher = {Preprints},
	journal = {Preprints},
	doi = {10.20944/preprints202512.1276.v1},
	url = {https://www.preprints.org/manuscript/202512.1276/v1}
}

@article{ABrain.NN-RAG,
	title={A Retrieval-Augmented Generation Approach to Extracting Algorithmic Logic from Neural Networks},
 	author={Khalid, Waleed and Ignatov, Dmitry and Timofte, Radu},
	journal = {arXiv preprint},
  	volume  = {arXiv:2512.04329},
  	url = {https://arxiv.org/abs/2512.04329},
 	year={2025}
}

@article{Gado2025llm,
	title={{VIST-GPT}: Ushering in the Era of Visual Storytelling with {LLMs}?},
	author={Gado, Mohamed and Taliee, Towhid and Memon, Muhammad Danish and Ignatov, Dmitry and Timofte, Radu},
	journal = {arXiv preprint},
  	volume  = {arXiv:2504.19267},
  	url = {https://arxiv.org/abs/2504.19267},
	year={2025}
}

@article{Aboudeshish2025augmentation,
	title={{AUGMENTGEST}: CAN RANDOM DATA CROPPING AUGMENTATION BOOST GESTURE RECOGNITION PERFORMANCE?},
	author={Aboudeshish, Nada and Ignatov, Dmitry and Timofte, Radu},
	journal = {arXiv preprint},
	volume  = {arXiv:2506.07216},
	url = {https://arxiv.org/abs/2506.07216},
	year={2025}
}

@InProceedings{ABrain.Prompt,
	title = {Enhancing {LLM}-Based Neural Network Generation: Few-Shot Prompting and Efficient Validation for Automated Architecture Design},
	author = {Raghuvir Duvvuri and Chandini Vysyaraju and Avi Goyal and Dmitry Ignatov and Radu Timofte},
	booktitle={Proceedings of the IEEE/CVF Conference on Computer Vision and Pattern Recognition Workshops (CVPRW)},	
	year={2026},
         note={to appear}
}

@InProceedings{ABrain.CV_Channel,
	title={Closed-Loop {LLM} Discovery of Non-Standard Channel Priors in Vision Models},
 	author={Uzun, Tolgay Atinc and Ignatov, Dmitry and Timofte, Radu},
	booktitle={Proceedings of the International Conference on Pattern Recognition (ICPR)},	
	year={2026},
         note={to appear}
}

@article{ABrain.Feedback_Memory,
	title={Resource-Efficient Iterative {LLM}-Based {NAS} with Feedback Memory},
	author={Gu, Xiaojie  and Ignatov, Dmitry and Timofte, Radu},
	journal = {arXiv preprint},
	volume  = {arXiv:2603.12091},
	url = {https://arxiv.org/abs/2603.12091},
	year={2026}
}

@InProceedings{ABrain.MobileAgeNet,
	title = {{MobileAgeNet}: Lightweight Facial Age Estimation for Mobile Deployment},
	author = {Arun Kumar and Aswathy Baiju and Radu Timofte and Dmitry Ignatov},
	booktitle={Proceedings of the IEEE/CVF Conference on Computer Vision and Pattern Recognition Workshops (CVPRW)},	
	year={2026},
         note={to appear}
}

@InProceedings{ABrain.MobileDenoising,
	title = {Real Image Denoising with Knowledge Distillation for High-Performance Mobile {NPUs}},
	author = {Faraz Kayani and Sarmad Kayani and Asad Ahmed and Radu Timofte and Dmitry Ignatov},
	booktitle={Proceedings of the IEEE/CVF Conference on Computer Vision and Pattern Recognition Workshops (CVPRW)},	
	year={2026},
         note={to appear}
}

@article{ABrain.Delta,
  author  = {Santosh Premi Adhikari and Radu Timofte and Dmitry Ignatov},
  title   = {Delta-based neural architecture search: {LLM} fine-tuning via code diffs},
  journal = {arXiv preprint},
  volume  = {arXiv:2605.04903},  
  year    = {2026}
}
	}
	\end{document}